\documentclass[10pt,a4paper]{article}

\usepackage{cogsci}
\usepackage{pslatex}
\usepackage{apacite}
\usepackage{graphicx}
\usepackage{amssymb}
\usepackage{amsmath}
\usepackage{subcaption}  
\usepackage{float} 
\usepackage{algorithm} 
\usepackage{algpseudocode} 
\usepackage{comment}

\title{Emergent social transmission of model-based representations without inference}

\author{\textbf{Silja Ke\ss ler$^{1}$}(silja.kessler@student.uni-tuebingen.de), \textbf{Miriam Bautista-Salinero$^{1}$}, \textbf{Claudio Tennie$^{1}$}\\\& \textbf{Charley M. Wu$^{2,3}$ (charley.wu@tu-darmstadt.de)} \\
$^1$ University of T\"ubingen, T\"ubingen, Germany\\
$^2$ Technical University Darmstadt, Darmstadt, Germany\\
$^3$ Hessian AI, Darmstadt, Germany
}

\begin{document}

\maketitle

\begin{abstract}
How do people acquire rich, flexible knowledge about their environment from others despite limited cognitive capacity? Humans are often thought to rely on computationally costly mentalizing, such as inferring others’ beliefs. In contrast, cultural evolution emphasizes that behavioral transmission can be supported by simple social cues. Using reinforcement learning simulations, we show how minimal social learning can indirectly transmit higher-level representations. We simulate a naïve agent searching for rewards in a reconfigurable environment, learning either alone or by observing an expert---crucially, without inferring mental states. Instead, the learner  heuristically selects actions or boosts value representations based on observed actions. 
Our results demonstrate that these cues bias the learner's experience, causing its representation to converge toward the expert’s. Model-based learners benefit most from social exposure, showing faster learning and more expert-like representations. These findings show how cultural transmission can arise from simple, non-mentalizing processes exploiting asocial learning mechanisms.

\end{abstract}

\section{Introduction}
Reinforcement learning (RL) offers a powerful framework for understanding how agents acquire knowledge through interaction with their environment \cite{sutton2018reinforcement}. Traditionally, this framework has focused on asocial learning: individual agents learn by interacting with their environment and updating internal representations based on experience. While an asocial framing has yielded substantial advances \cite{shteingart2014reinforcement, o2015structure}, it stands in contrast to how humans and other animals leverage \textit{social} information \cite{Laland2004-pq, olsson2020neural} to accelerate learning \cite{rendell2010copy, park2017integration}, reduce exploration costs \cite{witt2024flexible, garg2022individual}, and improve performance in complex environments \cite{wu2025adaptive, hawkins2023flexible}. 

Thus, recent work in RL has increasingly incorporated social learning mechanisms, to both understand human social cognition \cite{wu2025adaptive, hackel2020reinforcement, najar2020actions} and for engineering applications where learning from the environment is costly \cite{prasad2024moveint, urain2025survey}. Learning from others is often modeled as \textit{mentalizing} \cite{Baker_Saxe_Tenenbaum_2009, Jara-Ettinger_2019}---inferring latent mental states such as goals, preferences, or beliefs about the structure of the environment from observed actions \cite<also known as Theory of Mind;>[]{tomasello_cultural_1999}. Techniques like inverse reinforcement learning  \cite<IRL;>[]{Russell_1998, Jara-Ettinger_2019, jern2017people, collette2017neural} formalize this process by inverting the RL model to ``unpack'' observed behavior into underlying value or model-based representations of the world. Although powerful, such social inference methods are computationally costly and intractable in many settings \cite{wu2022representational}.

In contrast, cultural evolution research has long emphasized how sophisticated patterns of behavioral transmission can emerge from low-cost social cues \cite{heyes2018cognitive, mesoudi2016cultural}. Individuals often rely on simple heuristics, such as selectively imitating successful \cite{mcelreath2008beyond} or prestigious individuals \cite{jimenez2019prestige, boyd1988culture} without needing to infer mental states. 
These mechanisms can nonetheless support the accumulation of complex behaviors across generations \cite{legare2015imitation, rendell2010rogers}, suggesting an alternative pathway for effective social learning without the costly inference required for mentalizing \cite{roberts2025environmental, uchiyama2023model}. 

Here, we explore how higher-level representations of value and model-based state transitions can emerge in RL agents from simple forms of social learning---without explicit mentalizing---and how they support generalization to novel conditions. We simulate agents that search for rewards in a spatially reconfigurable environment, learning either independently or by observing an expert. Crucially, learners do not attempt to infer the expert’s goals or internal model of the environment. Instead, they rely on simple social cues, adopting strategies more similar to imitation \cite{byrne_learning_1998} or local enhancement \cite{Galef1988-bc} than mentalizing.
Specifically, we consider two strategies at different representational levels: \textit{decision biasing} \cite<DB;>[]{toyokawa_social_2019}, which biases the agent’s policy, and \textit{value shaping} \cite<VS;>[]{najar2020actions}, which shapes the agent's expected values for actions based on the expert's behavior.
Both strategies are applied to model-free and model-based RL agents, either guiding learning either through model-free habit formation or via model-based planning \cite{drummond2020model, hackel2019model}.

Across experiments, our results show that social learners---particularly those with model-based architectures---adopt more expert-like value and belief representations, supporting effective generalization to novel reward configurations and starting positions. These results demonstrate that simple, non-mentalizing social mechanisms can produce higher-level, flexible representations, highlighting a computational pathway through which complex cultural transmission can emerge from minimal social information.

\section{Methods}
\begin{figure}
    \centering
    \includegraphics[width=1\linewidth] {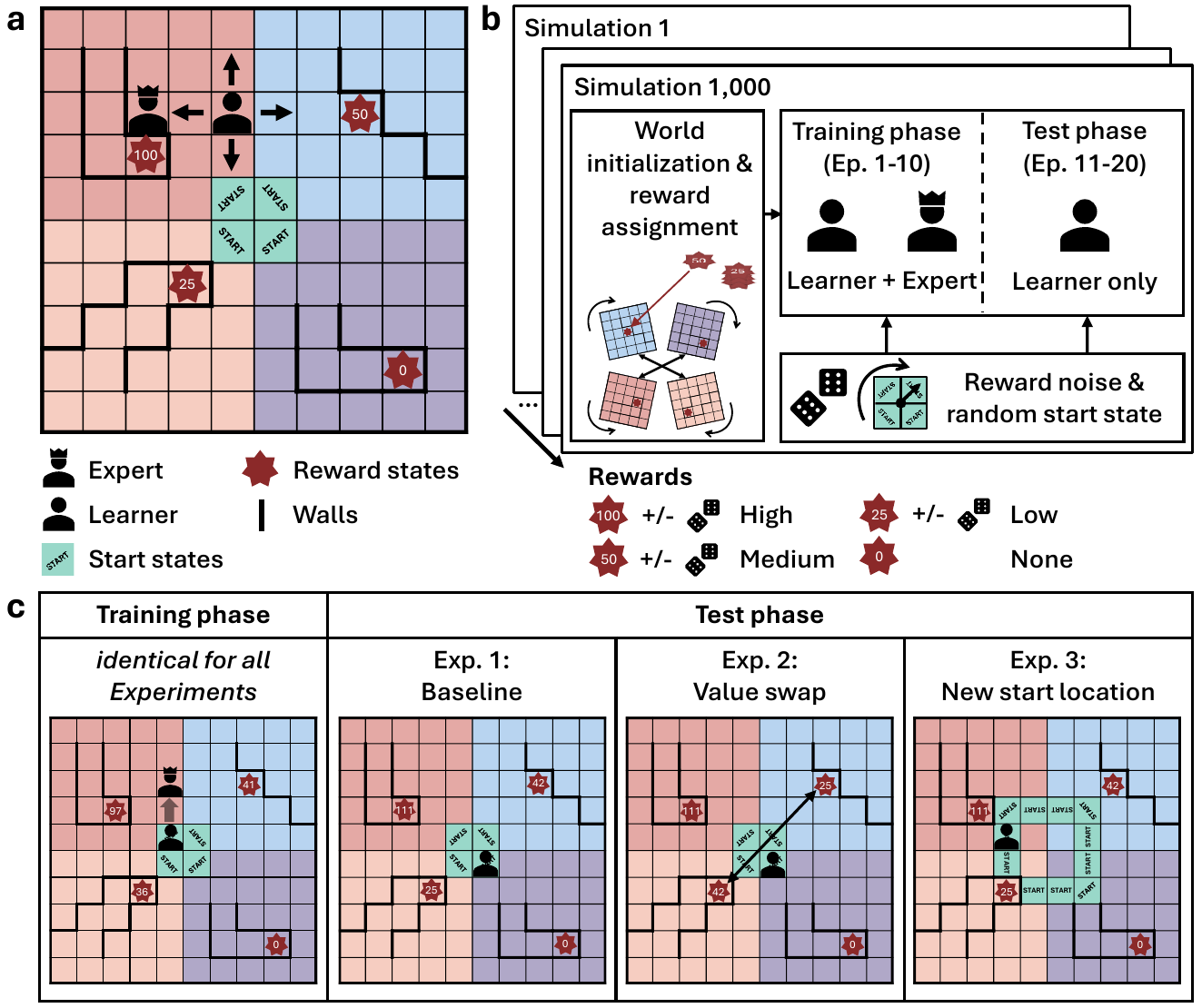}
    \caption{\textbf{Simulations}. \textbf{a}) Grid-world environment composed of four quadrants with fixed walls and designated reward states. \textbf{b}) For each simulation, quadrants were randomly rotated and arranged, while rewards were randomly assigned to designated reward states. The simulations comprised a training phase with an observable expert agent (over-trained model-based RL), followed by a test phase without the expert. A different start position was randomly selected on each episode (from the four central states), and stochastic noise was added to reward observations (see Methods). \textbf{c}) Experimental manipulations applied during the test phase: no change (Exp. 1), swap of two randomly selected rewards (Exp. 2), or shifted start positions (Exp. 3).}
    \vspace{-1.5em}
    \label{fig1}
\end{figure}

We simulated agent behavior across three experiments ($n=1,000$ per experiment) using a spatially reconfigurable environment, resembling a foraging-themed board game. Agents navigated generated grid-worlds over 20 episodes to acquire rewards. During training, agents learned through a combination of individual experience and observing an expert. Agent performance was then evaluated in the test phase, in which all agents continued to learn solely through individual experience, to assess generalization in the absence of the expert. 

\noindent \textbf{Environment and Rewards.}
The environment was composed of four quadrants, each forming a $5\times5$ grid of discrete states.
Each quadrant contained a fixed configuration of walls blocking movement and a single designated reward state (Fig.~\ref{fig1}a).
In each simulation, we rotated and rearranged these quadrants to 
generate $4!\cdot4^4=6,144$ unique configurations of $10\times10$ grid worlds.
Reward values for each designated reward state were randomly assigned (without replacement) at the start of each simulation from four values: 0 (no reward), 25 (low), 50 (medium), and 75 (high). 
To introduce stochasticity, noise $\epsilon$ was applied independently to each reward observation using a shifted binomial distribution $\epsilon \sim \mathrm{Binomial}(n = 400, p = 0.5) - 200$, yielding integer values with a mean of 0 and a variance of 100. 

\noindent \textbf{Procedure.}
At the start of each episode, the agent was placed in one of four start states at the center of the board. The agent moved between adjacent states by executing actions (up, down, left, right), freely moving between quadrants,  constrained by walls. Each attempts to move into a wall resulted in the agent remaining in the prior state. Actions not yielding a positive reward incurred a cost (negative reward) of $-1$. An episode terminated either when a positive reward was obtained or after 40 actions.

Simulations were divided into a \textit{training phase} (episodes 1-10) and a \textit{test phase} (episodes 11-20; Fig.~\ref{fig1}b). During the training phase, learners were accompanied by an expert, which was an asocial model-based RL agent (see Fig.~\ref{fig2}) that was pre-trained for 120 episodes in the same environment. Thus, the training phase offered social learning agents (defined below) the ability to combine information from the expert with knowledge gained through their own experiences. In contrast, asocial learners could only acquire knowledge through direct interactions with the environment. In the subsequent test phase, the expert was removed, meaning all learners could only learn asocially. 

\noindent \textbf{Experiments.}
We conducted three experiments to assess the learners’ ability to generalize knowledge acquired during training (Fig.~\ref{fig1}c). Exp.~1 established baseline performance and examined the transfer of values and model-based beliefs from the expert to social learners, without any changes to the environment. In Exps.~2 and 3, the test phase introduced additional modifications to the task to assess the learner's robustness to changes in the environment. Exp.~2 modified the reward structure, by swapping the values of two randomly selected reward states. Exp.~3 modified the starting locations, by shifting start states one tile towards the corners (excluding reward states), while keeping the spatial layout fixed.

\subsection{Computational models}
\begin{figure}
    \centering
    \includegraphics[width=1\linewidth]{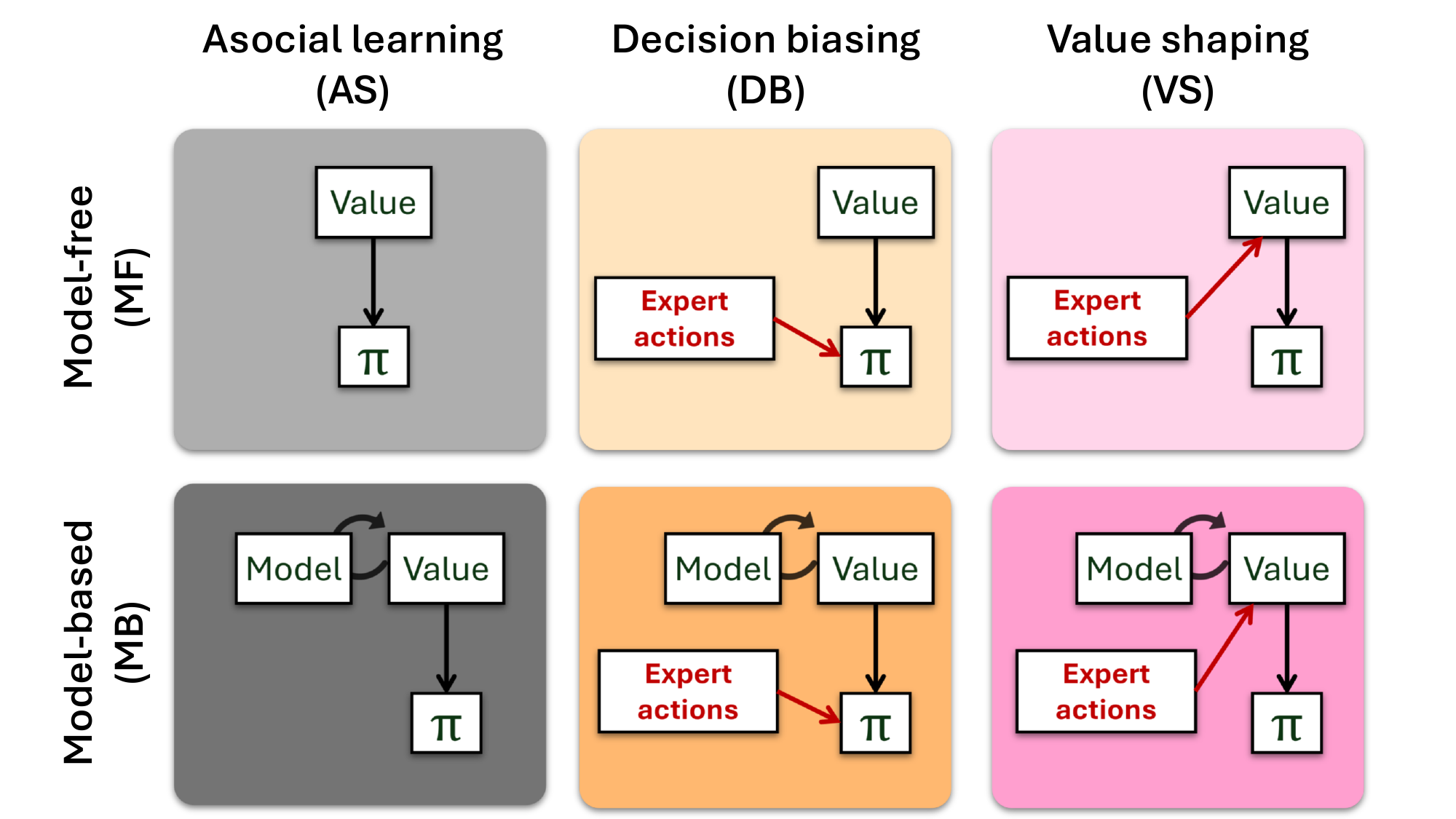}
    \caption{\textbf{Learning Strategies.} Six learning conditions were tested, combining model-free (MF) and model-based (MB) RL with either asocial learning (AS) or social learning from an expert. Social learning was implemented at two levels: policy-based (DB) and value-based (VS).}
    \label{fig2}
    \vspace{-1em}
\end{figure}

We compared six computational learning models (Fig.~\ref{fig2}) with a $2 \times 3$ factorial design contrasting model-free vs model-based RL across three forms of social learning:  asocial learning (AS), decision biasing (DB), and value shaping (VS). Asocial learners exclusively relied on individual experiences, whereas social learners additionally incorporated expert information either at the action selection level by biasing their policy toward the expert’s behavior (DB), or at the value learning level by directly shaping value representations (VS) based on observed expert actions.

\subsubsection{Model-free (MF) learning} was implemented as Q-learning with Temporal Difference (TD) prediction updates:
\begin{equation}
    Q(s,a) \leftarrow Q(s,a) + \alpha\left(R(s) + \gamma \max_{a'} Q(s',a') - Q(s,a)\right)
\end{equation}
\noindent where $Q(s,a)$ is the state-action value indicating the expected cumulative rewards for performing action $a$ in state $s$. Q-values were initialized uniformly to 1 for all state–action pairs and updated based on the learning rate $\alpha$ and proportional to the difference between the prior Q-value and the sum of the reward obtained in the current state $R(s)$ and the maximum next-state Q-value $\max_{a'} Q(s',a')$ discounted by a temporal discount factor $\gamma \in [0,1]$.

Using these learned Q-values, the asocial learner computed a policy $\pi_{asoc}$ based on a softmax function whose randomness was controlled by the inverse-temperature parameter $\beta$:
\begin{equation}
\pi_{asoc}(a|s) \propto \exp\left(\beta \cdot Q(s,a)\right)
\label{eq:pi_asoci}
\end{equation}
\subsubsection{Model-based (MB) learning} was implemented using Dyna-Q \cite{Sutton_1990}, a variant of Q-learning in which the agent maintains an internal model (i.e., beliefs) of the environment, defined as  the transition probabilities between states given an action, $B(s'|s,a)$. 
Beliefs were updated  via the Delta-rule \cite{antonov2025exploring}:
\begin{equation}
B(s'|s,a) \leftarrow B(s'|s,a) +  \eta\left(\delta(s',s) -  B(s'|s,a) \right)
\end{equation}

\noindent where the Kronecker delta $\delta(s',s)=1$ if the transition $s\rightarrow s'$ is successful, and 0 otherwise. 

These beliefs could be used to update Q-values more efficiently through simulated experiences (dependent on $B$), rather than solely relying on direct RL by executing actions in the world. This supports planning-based decision-making and is a cornerstone of modern theories of human learning \cite{daw2011model, drummond2020model, wu2022representational}. The following model-based planning loop was executed $k$ times after each action, with  $k \sim \textrm{Pois}(\lambda)$ sampled each time from a Poisson distribution to introduce stochasticity and $\lambda$ is treated as a free parameter (planning rate). Q-values and beliefs were maintained across episodes, such that knowledge acquired during training carried over into the test phase.
\begin{algorithm}
\small
	\caption{Dyna-Q inner loop} 
	\begin{algorithmic}[1]
		\For {$i=1,2,\ldots,k$}
				\State $S \leftarrow$ random previously observed state
                \State $A \leftarrow$ random action previously taken in $S$
				\State $R,S' \leftarrow B(S,A)$ 
    			\State  $Q(s,a) \leftarrow Q(s,a) + \alpha\left(R(s) + \gamma \max_{a'} Q(s',a') - Q(s,a)\right)$ 
		\EndFor \label{alg:dynaQ}
	\end{algorithmic} 
\end{algorithm}
\vspace{-1em}

\begin{figure*}[t]
    \centering
    \includegraphics[width=\textwidth]{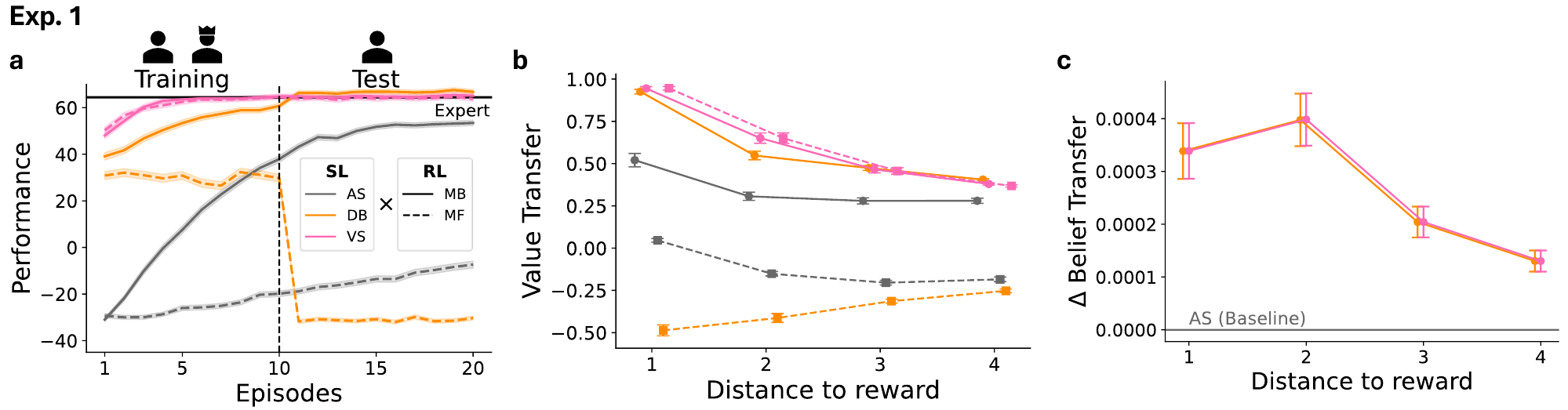}
    \vspace{-2em}
    \caption{\textbf{Experiment 1.} \textbf{a}) Performance: mean cumulative reward for different learning strategies combining (a-)social learning (SL) and reinforcement learning (RL) strategies. The vertical dotted line represents the transition between training and test phase. The black horizontal line shows the expert's mean performance over episodes observed by the social learners. \textbf{b}) Value transfer: mean correlation between learner's and expert's value function, grouped by distance to reward states. \textbf{c}) Belief transfer: mean correlation between learner and expert transition matrices, grouped by distance to reward states. Correlations are shown for model-based social learners normalized relative to the asocial model-based learner (horizontal grey line). Shaded areas and error bars denote the standard error of the mean (SEM) across simulations.}
    \label{fig3}
    \vspace{-1em}
\end{figure*}

\subsubsection{Decision-biasing (DB)} was inspired by algorithmic accounts of social learning in bandit tasks \cite{toyokawa_social_2019, witt2024flexible}, but modified for our sequential navigation task. We implemented DB by combining the asocial policy (Eq.~\ref{eq:pi_asoci}) with a purely social policy $\pi_{soc}$ during training, weighted by mixture parameter $\omega \in [0,1]$:
\begin{equation}
\pi_{\textrm{DB}} = (1-\omega)\pi_{asoc}(s,a)  + \omega\pi_{soc}(s,a)  
\label{eq:db}
\end{equation}
The social policy $\pi_{soc}$ deterministically selects the action that minimizes the distance to expert:
\begin{equation}
\pi_{\text{soc}}(a \mid s)
= \mathbb{I}\!\left[
a = \arg\min_{a'} D\bigl(s^{t}_{\text{expert}}, s'(s,a')\bigr)
\right]
\label{eq:pi-soc}
\end{equation}
\noindent where the indicator function $\mathbb{I}[\cdot]$ assigns 1 to the action that minimizes $D$, and 0 otherwise. Distance $D$ corresponds to the expected number of actions needed to reach the expert's last observed state $s_{expert}^{t}$ from the learner's current state $s'$. For MF agents, $D$ is approximated using Manhattan distance (ignoring boundaries or walls, since MF agents do not represent environmental structure). For MB agents, $D$ is computed via value iteration under the agent's beliefs: $D(s) = \min_{a}\sum_{s'}B(s'|s,a')(1 + D(s'))$.
When the expert terminated prior to the learner, the expert's last visited location was frozen.

\subsubsection{Value shaping (VS)} was inspired by theoretical accounts of the local enhancement of stimuli based on observing others engaging with them \cite{Spence1937-gr, Galef1988-bc}. Following \citeA{najar2020actions}, we implemented VS by augmenting the learner’s value representations during training for actions performed by the expert in a given state:
\begin{equation}
Q(s_{expert},a_{expert})\leftarrow Q(s_{expert},a_{expert}) + \kappa\mathbb{I}(s_{expert},a_{expert}).
\label{eq:vs}
\end{equation}
\noindent $\kappa$ is a bonus parameter that directly shapes the learner's value function by enhancing state-action pairs observed in the expert.
The boosted Q-values were then passed through a softmax (cf. Eq.~\ref{eq:pi_asoci}), increasing the likelihood of selecting actions observed in the expert in the corresponding state.
Once the expert reached a termination step, the VS learner reduced to asocial learning for the remainder of the episode. 

\subsubsection{Parameter optimization.}
We optimized each model’s hyperparameters using the differential evolution algorithm \cite{storn1997differential} to maximize performance (cumulative reward) across 1,000 independent simulations in the baseline environment. Optimization for social learners was based solely on the 10 training episodes, whereas asocial agents and the expert were optimized using all available episodes (20 and 120, respectively).
Learning rates $\alpha$ and $\eta$ for social learners were fixed to the values optimized for asocial learners, due to joint optimization consistently converging to low learning rates that suppressed individual learning and led to poor test performance. Candidate parameters were optimized in a transformed, unbounded space to enforce valid ranges (e.g., log-transform for $\beta \in [0,\infty])$ and a fixed random seed was used to ensure reproducibility.

\section{Results}
We first evaluated the performance of all agents across episodes, distinguishing between training and test phases to characterize how effectively agents acquired and applied knowledge, depending on their learning strategy. Next, we examined how information was indirectly transferred from expert to social learners by analyzing value and belief representations, assessing whether social learners acquired similar value and state transition representations as encoded by the expert. Finally, we tested the robustness of the learned knowledge under environmental changes: in Experiment 2 we assessed how well learners generalized to modified reward configurations, and in Experiment 3 we modified the starting position to evaluate whether the belief transferred from the expert was preserved and continued to support behavior under novel start conditions. Across analyses, variability across simulations was quantified using the standard error of the mean (SEM).

\subsection{Exp. 1: Value/belief transfer without mentalizing}
Exp.~1 established a baseline in which training and test phases were identical except for the expert's removal. Performance was quantified as the mean cumulative reward per learner across simulations, demonstrating how effectively different learning strategies enabled agents to navigate the environment and collect rewards over the course of the 20 episodes. 

Starting with the asocial agents (AS), we observed a strong advantage of model-based (MB) over model-free (MF) learning (Fig. \ref{fig3}a). However, all social learning agents (DB and VS) outperformed the best asocial agent (AS-MB) in the training phase by leveraging information from the expert. Among social strategies, VS agents were superior, reaching expert-level performance with no difference between MB and MF. This reflects differences in how social information is incorporated. VS agents directly learn state-action values from the observed expert behavior, enabling effective generalization across the state space, whereas DB agents acquire values indirectly from expert trajectories, which confines learning signals to a narrow region and reduces flexibility. Moreover, combining social and individual information at the policy level can create conflict between following the expert and exploiting the DB agent's own experience.
After the removal of the expert in the test phase, VS performance remained stable, demonstrating the robustness of value-based social learning---notably, also for MF. In contrast, DB agents diverged sharply: DB-MF performance collapsed below AS-MF, suggesting an over-reliance on the expert that produced an unreliable value function beyond the expert’s trajectories. DB-MB, however, continued to improve and ultimately surpassed both VS agents, replanning more effectively once the bias was removed.

We then computed indirect \textit{value transfer} based on the correspondence between the learner’s and expert’s state-action value representations at the end of each simulation (averaged across simulations). Specifically, we computed Spearman correlations between value functions across states, grouped by distance from the nearest reward to account for goal-relevance (i.e., states closer to rewards have greater relevance).  This assessed how effectively the learner reconstructs the expert's value information. Figure~\ref{fig3}b shows that except for DB-MF, all social agents achieved greater value transfer compared to asocial baselines (AS-MB and AS-MF). While asocial agents were bound to acquire similar value representations as the expert due to interacting with the same environment, social agents (except for DB-MF) achieved a substantial boost, providing a mechanistic explanation for their better performance. This value transfer effect also displayed goal-relevance, with stronger transfer for states closer to rewards. 

Lastly, we quantified indirect \textit{belief transfer} for MB learners as the correspondence between learner and expert transition matrices at the end of each simulation. We computed Pearson correlations across states, grouped by distance from rewards, and normalized them relative to AS-MB as an asocial baseline (Fig. \ref{fig3}c). We observed positive belief transfer for both MB social learners (equivalent for DB-MB and VS-MB), indicating that they acquired an internal model of state transitions that more closely matches the expert’s compared to the asocial baseline. This effect was also influenced by goal-relevance, with states closer to rewards corresponding to greater belief transfer to our model-based social learners.

\subsection{Exp. 2: Robustness to changes in reward}
In Exp.~2, the test phase additionally included swapping the values of two randomly selected reward states (Fig. \ref{fig1}c). As in Exp.~1, we quantified performance as the mean cumulative reward per learner across simulations to assess how effectively different learning strategies generalized to the altered reward contingencies, focusing on the test phase since the training phase was identical to Exp.~1.

Test performance initially dropped for all agents (except AS-MF), reflecting the challenge of adapting to new reward contingencies---specifically for VS agents whose boosted values were now potentially misleading. The most pronounced drop occurred for DB-MF, which showed a similar collapse as in Exp.~1.
However, all MB agents adapted to the new reward configuration and recovered performance.

To examine the underlying mechanisms, we analyzed \textit{value accuracy} by computing the Spearman correlation between each learner’s final value function (averaged across simulations) and the optimal value function for the new reward configuration, derived from the Bellman \citeyear{Bellman1957} equation: 
\begin{equation}
Q^*(s,a) = \sum_{s'} P(s' \mid s,a) \Big[ R(s,a,s') + \gamma \max_{a'} Q^*(s',a') \Big]
\label{eq:bellman}
\end{equation}
with $\gamma = 0.99$.
Here, we only included states visited at least once by the learner to avoid distortions from unvisited states. Correlations were again grouped by distance from the nearest reward to account for goal-relevance. 

\begin{figure}[t]
    \centering
    \includegraphics[width=1\linewidth]{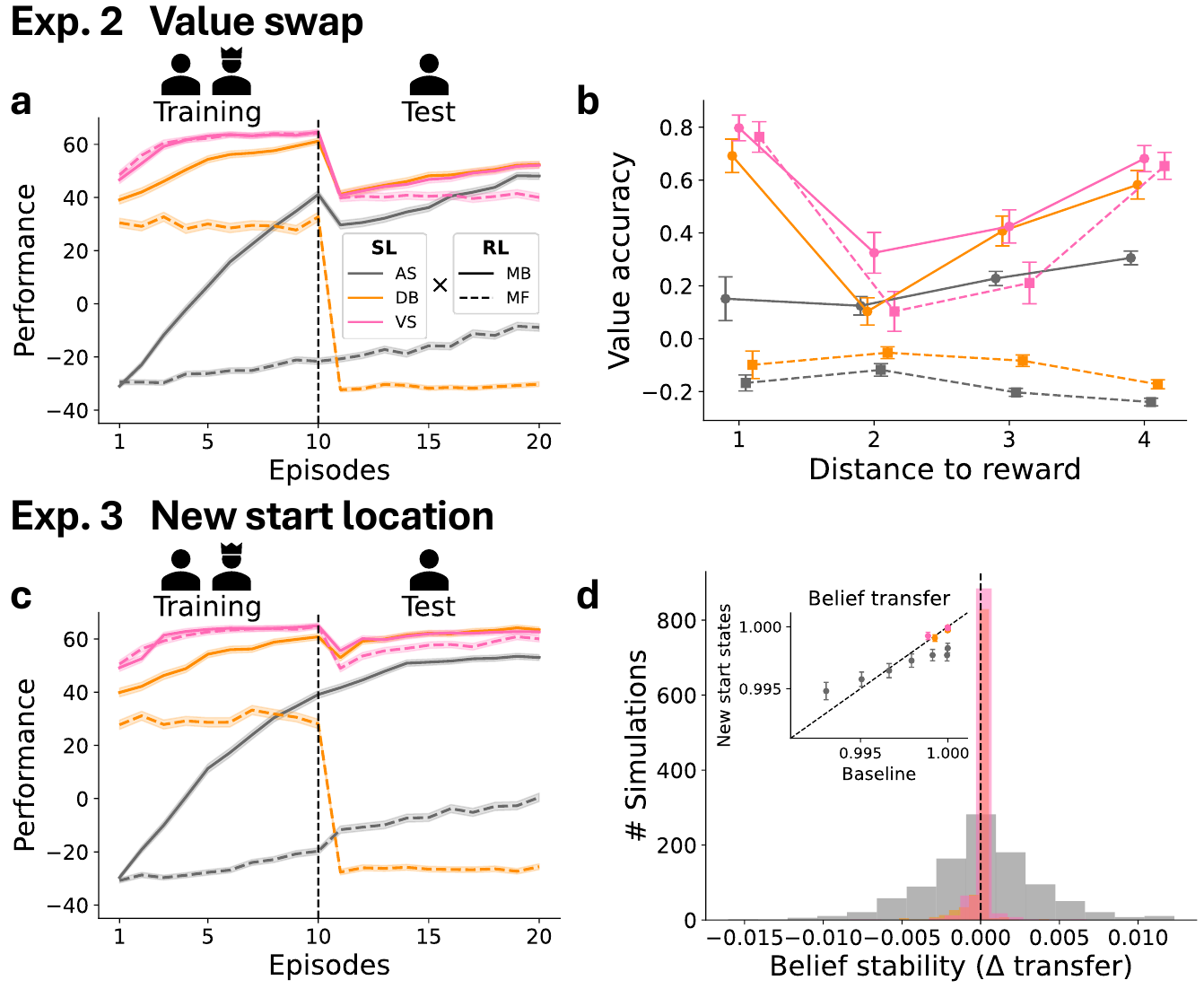}
    \vspace{-1.5em}
    \caption{\textbf{Exps.~2 and 3.} 
    \textbf{a}) Exp.~2 performance. \textbf{b}) Exp.~2 value accuracy: mean correlation between the learner's and the optimal value function for the modified reward structure (Eq.~\ref{eq:bellman}), grouped by distance to reward. 
    \textbf{c}) Exp.~3 performance. 
    \textbf{d}) Exp.~3 belief stability (MB only). Deviation in belief transfer between baseline and new start locations in the test phase across simulations; zero (dashed line) indicates robust preservation of model-based representations. Inset: comparison of belief transfer baseline (x-axis) vs. new start location (y-axis); points represent binned averages across simulations (quantiles) and error bars indicate 95\% confidence intervals. The dashed diagonal line shows $y=x$.
    }
    \label{fig4}
    \vspace{-1.5em}
\end{figure}

Our results show that social learning facilitates better generalization to changes in reward structure, with all social agents achieving greater value accuracy compared to asocial baselines (AS-MB and AS-MF), except for DB-MF, whose poorly aligned value representations explain its poor performance (Fig. \ref{fig4}b). Notably, accuracy dropped at $d=2$ for all agents except AS-MF and DB-MF, likely because one reward state was reachable via alternative paths diverging at that point (Fig. \ref{fig1}a). Further analyses indicated that agents consistently committed to one of the routes and as a result did not update the values of the unchosen, equally optimal path. AS-MF and DB-MF, in contrast, show a slight increase in accuracy at $d=2$. These agents were more exploratory and therefore acquired broader knowledge about regions surrounding rewards. However, they more frequently settled for suboptimal rewards, resulting in lower values for states directly preceding a reward ($d=1$). 

\subsection{Exp. 3: Robustness to changes in starting location}
In Exp.~3, we tested robustness to changes in the starting positions (Fig. \ref{fig1}c). 
Focusing on the test phase, asocial learners continued to perform well, showing little sensitivity to the new start positions. All social learners (except for DB-MB) experienced a small initial dip in performance, but rapidly recovered to training-level performance. DB-MF suffered again a severe collapse similar to Exps.~1 and ~2. These results reinforce the finding that policy-based social learning is fragile under environmental changes unless paired with a model-based system. Importantly, model-based social learners (DB-MB and VS-MB) maintained a clear performance advantage over the asocial baseline (AS-MB) throughout the test phase, highlighting the robustness of model-based social strategies.

To examine whether beliefs acquired by model-based agents via the expert's actions were preserved under novel starting positions (Fig.~\ref{fig4}d), we quantified belief stability as the deviation in indirect belief transfer (correlation between agent and expert transition matrices, grouped across distances) between Exp.~1 (baseline start positions) versus Exp.~3 (new start positions). Social learners, particularly VS-MB, showed notably less deviation than AS-MB, indicating that their reconstructed beliefs were largely preserved and remained highly similar to the expert (Fig.~\ref{fig4}d inset). This showcases that social learners acquired more accurate, robust and presumably more global beliefs by leveraging the expert’s knowledge---resulting in internal representations that generalize more effectively across starting positions. 

Overall, these results demonstrate that particularly model-based social learners not only acquire similar structural knowledge to the expert, but also exhibit improved performance that remains robust under environmental changes---all without mentalizing.

\section{Discussion}
We investigated how rich, flexible knowledge can be socially transmitted without explicit mentalizing, thus sidestepping costly inferences about others' goals or beliefs. Using simulations with both model-free (MF) and model-based (MB) reinforcement learning (RL) agents, we show that learners can acquire higher-level representations through simple social heuristics. Specifically, decision-biasing (DB) operates at the policy-level, minimizing the distance to an expert demonstrator (Eqs.~\ref{eq:pi-soc}-\ref{eq:db}), while value shaping (VS) adds a value bonus to state-action pairs observed in the expert, akin to local enhancement (Eq.~\ref{eq:vs}).
Across three sets of simulations in a spatial foraging task, we found that social learning particularly benefited MB agents, which acquired more accurate and robust representations, while also generalizing better to changes in rewards and starting locations.

What mechanisms underlie these results? Even simple social heuristics bias the learner's experience, which, when leveraged by model-based planning, can support more accurate representations of value and environment structure \cite{wu2022representational, uchiyama2023model}. In our simulations, MB agents use Dyna (Alg.~1) to simulate experiences and propagate values, analogous to hippocampal replay \cite{olafsdottir2018role, miller2017dorsal, vikbladh2019hippocampal}. This suggests that insight can emerge from naïve imitation \cite{lyons2007hidden} by exploiting planning mechanisms. This account is consistent with neural evidence that social observations evoke replay \cite{mou2025observational, clein2025representations}, pointing to a shared neural basis for social and asocial model-based learning.

Across both MB and MF agents, value-based social learning (VS) generally outperformed policy-based methods (DB), echoing a general principle in RL that value is a more generalizable representation than actions \cite{lehnert2020reward, wu2022representational}. Notably, VS maintained strong performance even under model-free learning (VS-MF). In contrast, policy-based social learning was fragile under model-free learning (DB-MF), but robust when paired with model-based mechanisms (DB-MB). A likely explanation is that DB agents overfit to the expert’s demonstrated trajectories, resulting in limited exploration of the state space. When the expert is removed, DB-MF agents are left with an underdeveloped value function and no mechanism to recover. DB-MB agents, however, can use their learned model to replan and update their value representation effectively.

These findings offer a potential perspective on cross-species differences between humans and non-human primates \cite<NHP;>[]{bandini2023naive, call2005copying}. Humans over-imitate, copying even irrelevant actions \cite{horner2005causal, lyons2007hidden}, whereas chimpanzees tend to emulate goals rather than copy specific behavior \cite{tennie2010evidence}. While the literature on model-based learning in NHPs is scarce \cite<but see>[for a task disincentivizing MB learning]{sato2023state}, differences in reliance on model-based planning may shift the cost–benefit trade-off between policy-based and value-based social learning \cite{wu2022representational}, which could explain why over-imitation is more adaptive in humans than in chimpanzees. 

Despite promising results, there are several limitations that could be addressed in future work. 
First, while we evaluated robustness to changes in rewards and starting locations, we did not manipulate the underlying environmental structure itself (e.g., changes to transition dynamics or wall layouts), which would allow us to better assess whether non-mentalizing social learning also supports the transfer of deeper structural knowledge \cite{ho2022planning}. 
Second, we did not incorporate mentalizing agents, e.g., agents using inverse reinforcement learning \cite<IRL;>[]{Russell_1998} to infer the expert's values or beliefs, into our simulation. Including such agents would provide an important comparative baseline, enabling more direct assessment of trade-offs between computationally costly mentalizing and simpler social heuristics like DB and VS \cite{devaine2014theory}.
Third, we modeled the expert as a passive demonstrator rather than providing pedagogical demonstrations \cite{velez2023teachers}. Future work could test how such demonstrations might interact with or amplify our minimal social learning mechanisms. Finally, our results are based on simulations, whereas testing these predictions in human behavioral experiments will be essential to assess whether similar non-mentalizing processes can also support higher-order social transmission in people.

In sum, we provide a computational account of how the indirect transmission of high-level representations can arise from minimal, non-mentalizing forms of social learning that exploit asocial model-based RL mechanisms.

\section{Acknowledgments}

We thank Ryutaru Uchiyama and members of the HMC Lab for helpful discussions in the development of this project. This work is supported by the European Research Council (ERC) under the European Union’s Horizon 2020 research and innovation programme ($C^4$: 101164709), the Hessian research funding programme LOEWE/4b//519/05/01.002(0022)/119, the Deutsche Forschungsgemeinschaft (German Research Foundation, DFG) under Germany’s Excellence Strategy (EXC 3066/1 ``The Adaptive Mind'', Project No. 533717223), and the Excellence Cluster ``Reasonable AI'' by the Deutsche Forschungsgemeinschaft (German Research Foundation, DFG) under Germany’s Excellence Strategy – EXC-3057.

\end{document}